\documentclass[runningheads]{llncs}
\usepackage{graphicx}
\usepackage[hidelinks]{hyperref}
\DeclareTextSymbolDefault{\dh}{T1}
\usepackage{url}
\usepackage{svg}
\graphicspath{ {./images/} }
\begin{document}
	\title{
		Improving Reinforcement Learning Efficiency with Auxiliary Tasks in Non-Visual Environments: A Comparison
		\thanks{Supported by the research training group ``Dataninja'' (Trustworthy AI for Seamless Problem Solving: Next Generation Intelligence Joins Robust Data Analysis) funded by the German federal state of North Rhine-Westphalia.}
	}
	\titlerunning{Improving RL Efficiency with Aux. Tasks in Non-Visual Environments}
	%
	\author{%
		Moritz Lange\inst{1}\orcidID{0000-0001-7109-7813}\and
		Noah Krystiniak\inst{1}\and
		Raphael C. Engelhardt\inst{2}\orcidID{0000-0003-1463-2706}\and
		Wolfgang Konen\inst{2}\orcidID{0000-0002-1343-4209}\and
		Laurenz Wiskott\inst{1}\orcidID{0000-0001-6237-740X}\\
	}
	\authorrunning{M. Lange et al.}
	%
	\institute{
		Institute for Neural Computation, Faculty of Computer Science, Ruhr-University Bochum, Bochum, Germany \\ \email{\{moritz.lange, noah.krystiniak, laurenz.wiskott\}@ini.rub.de} \and Cologne Institute of Computer Science, Faculty of Computer Science and Engineering Science, TH Köln, Gummersbach, Germany \\ \email{\{raphael.engelhardt, wolfgang.konen\}@th-koeln.de}
	}
	\maketitle              
	\begin{abstract}
		Real-world reinforcement learning (RL) environments, whether in robotics or industrial settings, often involve non-visual observations and require not only efficient but also reliable and thus interpretable and flexible RL approaches. To improve efficiency, agents that perform state representation learning with auxiliary tasks have been widely studied in visual observation contexts. However, for real-world problems, dedicated representation learning modules that are decoupled from RL agents are more suited to meet requirements. This study compares common auxiliary tasks based on, to the best of our knowledge, the only decoupled representation learning method for low-dimensional non-visual observations. We evaluate potential improvements in sample efficiency and returns for environments ranging from a simple pendulum to a complex simulated robotics task. Our findings show that representation learning with auxiliary tasks only provides performance gains in sufficiently complex environments and that learning environment dynamics is preferable to predicting rewards. These insights can inform future development of interpretable representation learning approaches for non-visual observations and advance the use of RL solutions in real-world scenarios.

	\end{abstract}

	\section{Introduction}
	
	In reinforcement learning (RL), the complex interplay of observations, actions, and rewards means that algorithms are often sample-inefficient or cannot solve problems altogether. State representation learning tackles this issue by making information encoded in observations, and possibly actions, more accessible. Mnih et al. \cite{mnih_playing_2013} were the first to introduce deep RL to extract information from the high-dimensional observations provided by Atari games. The neural networks of deep RL agents make it possible to implicitly extract representations of the input and thus enables the agent to  find a good policy.
	
	Munk et al. \cite{munk_learning_2016} additionally introduced predictive priors, learning targets that differ from the RL task but are also based on data generated by the environment. Other authors such as Legenstein et al. \cite{legenstein_reinforcement_2010}, Wahlstrom et al. \cite{wahlstrom_pixels_2015}, Anderson et al. \cite{anderson_faster_2015}, and Schelhamer et al. \cite{shelhamer_loss_2017} have proposed further learning targets and started to call these auxiliary tasks.
	
	Both Munk et al. \cite{munk_learning_2016} and Stooke et al. \cite{stooke_decoupling_2020} argue for decoupling representation learning with auxiliary tasks from solving the RL task of maximizing cumulative rewards. Approaches with separate representation learning modules are versatile as representations can replace raw observations and actions as inputs to arbitrary RL algorithms. The individual parts of such agents have distinct purposes; auxiliary tasks and RL task do not interfere with each other and representations are agnostic to the RL task. Segmenting systems into such distinct parts provides flexibility and aids interpretability as representations used as inputs become explicit rather than being hidden within layers inside networks. The distinction of integrated and decoupled representation learning however is not always easy, e.g. when a deep RL agent is simply furnished with additional prediction heads to enhance internal representations within its neural networks.
	
	Few works in RL learn interpretable representations (an exception is e.g. \cite{legenstein_reinforcement_2010}) and methods such as autoencoders, which can supposedly learn semantically meaningful representations (used for RL in \cite{van_hoof_stable_2016,wahlstrom_pixels_2015}), have been shown to be unreliable in this regard \cite{locatello_challenging_2019}.
	Another shortcoming of the field is that these and most other works study complex, visual environments such as Arcade games \cite{shelhamer_loss_2017} or race tracks \cite{de_bruin_integrating_2018}. Many RL problems, however, provide non-visual observations 
	that are often of lower dimensionality and do not have the same properties as visual data. Observations are thus not necessarily suited for methods designed for visual data, such as CNN-based autoencoders, and are also not as easily interpreted.
	In real-world applications, e.g. control problems such as system control in factory production lines, RL is underrepresented due to concerns about reliability and sample efficiency \cite{dulac-arnold_challenges_2019}. We believe that new methods for interpretable, modular representation learning will be required to change this.
	
	To the best of our knowledge, OFENet by Ota et al. \cite{ota_can_2020} is to date the only available method for decoupled representation learning with auxiliary tasks that works also with low-dimensional observations and has been used on non-visual environments. We use it in this paper to learn representations on different auxiliary tasks and compare them on non-visual environments of different complexity. Despite the fact that OFENet as a method does not produce interpretable representations, we are confident that our findings concerning auxiliary tasks will help researchers in developing new, interpretable methods for representation learning in RL.
	
	We conduct our comparison by investigating returns and sample efficiencies achieved with common auxiliary tasks on five diverse environments. These environments cover a range of observation and action dimensionalities, and varying levels of complexity in the relationship between observations, actions, and rewards. Decoupled representations are computed with OFENet and used as inputs to the off-policy RL algorithms TD3 and SAC \cite{fujimoto_addressing_2018,haarnoja_soft_2018}. Since one environment, FetchSlideDense-v1, cannot be solved with baseline TD3 or SAC, we additionally train agents with hindsight experience replay (HER, \cite{andrychowicz_hindsight_2017}).
	
	Our results show that representation learning with auxiliary tasks increases both maximum returns and sample efficiency for environments that are sufficiently complex and high-dimensional, but has little effect on simpler, smaller environments. We find that learning representations based on environment dynamics, for instance by predicting the next observation, is superior to using reward prediction. We discover that decoupled representation learning with an inverse dynamics task does not work with actor critic algorithms because gradients cannot be backpropagated. Interestingly, adding representation learning to TD3 makes it possible to train agents on the FetchSlideDense-v1 environment, even if baseline TD3 does not learn anything at all.
	
	\section{Related Work}
	
	Many works in recent years have made use of an auxiliary task to learn state representations. We cite multiple of these in Sect. \ref{sec:aux_tasks}. Ota et al. \cite{ota_can_2020} for instance, whose OFENet representation learning network we use here, predict the next observation from current observation and action. There are however few papers which compare auxiliary tasks to each other. Lesort et al. \mbox{\cite{lesort_state_2018}} have written a thorough survey of state representation learning, which summarizes different auxiliary tasks and includes a comprehensive list of publications. It is however a purely theoretical discussion of methods without any empirical comparisons or results. There are two empirical comparisons of auxiliary tasks (\cite{shelhamer_loss_2017,de_bruin_integrating_2018}), which differ in various aspects from ours. Shelhamer et al. \cite{shelhamer_loss_2017}, like us, compare auxiliary tasks on various environments. In contrast to us, they use Atari games with visual observations. Another difference is that they do not fully decouple representation learning from the RL algorithm. Instead, they merely connect a different prediction head to train the initial, convolutional part of the deep RL algorithm on auxiliary targets. Their results generally vary across environments. An interesting feature of their paper is the comparison of individual auxiliary tasks to a combination of several. Curiously, the combination never clearly outperforms the respective best individual tasks. The second comparison, by \mbox{de Bruin et al. \cite{de_bruin_integrating_2018}}, uses only one car race environment but with several race tracks. It provides multimodal observations which again include visual data. In contrast to the decoupled module we use, loss functions of auxiliary tasks and RL task are linearly combined, and auxiliary tasks are investigated by removing their individual loss terms from the combination. This means representation learning is inherently integrated into the RL agent and happens implicitly, rather than in a decoupled way such as in our work.

	\section{Auxiliary Tasks}
	\label{sec:aux_tasks}
	
\begin{figure}[h]
		\centering
		\includegraphics[width=1\textwidth]{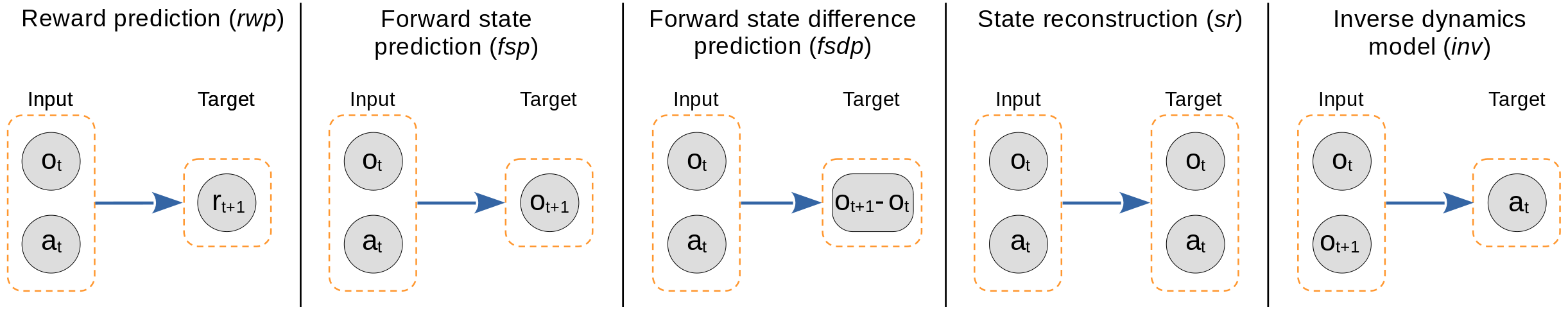}
		\caption{An overview of inputs and prediction targets of common auxiliary tasks.}
		\label{fig:overview_aux_tasks}
	\end{figure}

	In this section we present five common auxiliary tasks according to Lesort et al. \cite{lesort_state_2018}. Of these five, we will empirically compare three while the last two do not work with our setup. An overview of the tasks is presented in \mbox{Fig. \ref{fig:overview_aux_tasks}.} To discuss these tasks, we first need to briefly formalize the reinforcement learning problem: An environment provides reward $r_t$ and observation $o_t$ at time step $t$. The agent then performs an action $a_t$, which generates a reward $r_{t+1}$ and leads to the next observation $o_{t+1}$. This cycle is modeled by a Markov decision process, which means that there can be randomness in the transition from $o_t$ to $o_{t+1}$, given some $a_t$. The Markov property implies that $o_{t+1}$ only depends on $o_t$ and $a_t$ which already contain all past information. It does not depend on previous states or actions. The goal of the RL agent is to maximize cumulative expected reward (return). Altogether, these components are the ones available to auxiliary tasks, and various possible combinations are used.
	
	\textbf{Reward prediction (\textit{rwp})} is the task of predicting $r_{t+1}$ from $o_t$ and $a_t$. Works that use \textit{rwp} include \cite{munk_learning_2016,jaderberg_reinforcement_2016,shelhamer_loss_2017,oh_value_2017,hlynsson_reward_2021}. With decoupled representation learning, \textit{rwp} is limited in that it can only be applied to environments that provide non-trivial rewards, i.e. rewards that are not constant or sparse. Representations learned within the model might otherwise become decoupled from $o_t$ and $a_t$ since the model output $r_{t+1}$ is (nearly) independent of model inputs $o_t, a_t$. A RL algorithm trained on these representations could not learn anything at all. It can be argued that representations based on reward prediction have an advantage over those learned with other auxiliary tasks as they are optimized towards the actual RL task. On the other hand, \textit{rwp} is somewhat redundant to the RL task of maximizing returns, although it only considers the immediate next reward and is therefore less noisy \cite{shelhamer_loss_2017}.
	
	\textbf{Forward state prediction (\textit{fsp})} is the task of predicting $o_{t+1}$ from $o_t$ and $a_t$. It is a popular task and used e.g. in \cite{wahlstrom_pixels_2015,munk_learning_2016,van_hoof_stable_2016,pathak_curiosity-driven_2017,ota_can_2020}. In contrast to our work, several of these deal with high-dimensional image observations and therefore try to predict the next internal representation rather than the next observation in time. The \textit{fsp} task, as opposed to \textit{rwp}, can be applied to any kind of environment without conditions. Its task amounts to learning environment dynamics similar to e.g. in model-based RL. The \textit{fsp} task can thus be considered model-based RL, although we combine it with model-free RL algorithms.
	
	\textbf{Forward state difference prediction (\textit{fsdp})} describes the task of predicting $(o_{t+1}-o_t)$ from $o_t$ and $a_t$. The papers \cite{anderson_faster_2015,jaderberg_reinforcement_2016} use \textit{fsdp}. Conceptually, it is very similar to \textit{fsp}. While \textit{fsdp} also learns environment dynamics, Anderson et al. \cite{anderson_faster_2015} claim that successive observations are very similar and predicting only the difference thus gives more explicit insight into environment dynamics. The \textit{fsdp} task requires a notion of difference, though this is not a practical issue as observations are usually encoded as numerical vectors. In comparison with \textit{fsp}, the \textit{fsdp} task should provide an advantage in environments without excessive noise or significant changes between successive observations. That would mean \textit{fsp} is more robust, but \textit{fsdp} is particularly suited for environments simulating real-world physics.
	
	\textbf{State reconstruction (\textit{sr})} (used e.g. in \cite{jaderberg_reinforcement_2016,shelhamer_loss_2017}) is the task of reconstructing $o_t$ (and possibly $a_t$) from $o_t$ and $a_t$.  This is the classical autoencoder task, but does not make sense in our setup where data dimensionality is expanded by concatenation (see Sect. \ref{sec:ofenet}). Our representations thus always contain the raw $o_t$ and $a_t$, and reconstruction would amount to simply filtering these out. No useful representations could be learned. On a technical level, \textit{sr} is similar to \textit{fsp} but, crucially, does not learn environment dynamics. It therefore seems reasonable to assume that \textit{fsp} will in most cases be a better choice for learning representations for RL. The results of both \cite{jaderberg_reinforcement_2016} and \cite{shelhamer_loss_2017} confirm this.
	
\begin{figure}[h]
		\centering
		\includegraphics[width=0.8\textwidth]{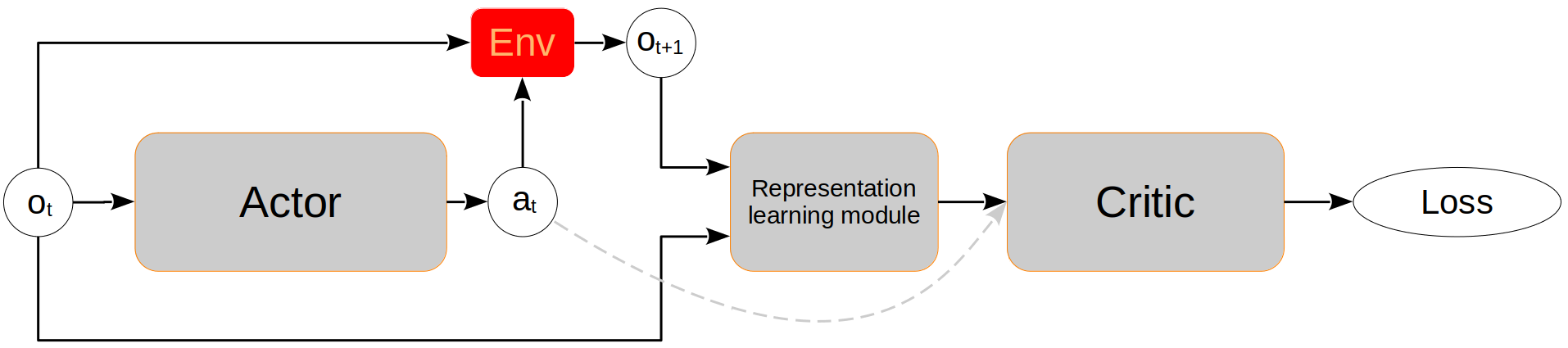}
		\caption{Diagram of information flow in an actor-critic setup with the \textit{inv} auxiliary task where the critic, or representation learning module, receives $o_t$ and $o_{t+1}$ (and potentially $a_t$) as input. If $o_{t+1}$ is part of the input to the critic, directly or through the module, the gradient of the critic loss cannot be propagated back to the actor as long as the environment (red) is not differentiable. Even if the action is additionally passed into the critic directly (dashed grey line) the actor will not get the true gradient.}
		\label{fig:inv_backprop_problem}
	\end{figure}

	The \textbf{inverse dynamics model (\textit{inv})}, framed as a learning task, predicts $a_t$ from $o_t$ and $o_{t+1}$. Works using this task include \cite{shelhamer_loss_2017,pathak_curiosity-driven_2017}. While \textit{fsp} and \textit{fsdp} focus on learning transition probabilities of the environment, the \textit{inv} task considers how actions of the agent affect changes in the environment. The \textit{inv} task does not work with actor-critic algorithms; using its representation as input to the critic renders the actor untrainable. Gradients would have to pass through the -- usually not differentiable -- environment in order to be propagated back from critic to actor. Figure \ref{fig:inv_backprop_problem} provides a visualization of this problem.
	
	Various \textbf{other priors} have been proposed by different authors (for a list, see~\cite{lesort_state_2018}). Noteworthy examples include the slowness principle \cite{legenstein_reinforcement_2010} and the robotics prior \cite{jonschkowski_learning_2015}. However, these are not as commonly used as the tasks above, and many are even problem-specific. We thus exclude them from our comparison.
	
	In addition to the tasks above, there are several works on combining auxiliary tasks. A popular combination is \textit{fsp} or \textit{fsdp} with \textit{rwp} (e.g. \cite{munk_learning_2016,jaderberg_reinforcement_2016}), various others exist. Lin et al. \cite{lin_adaptive_2019} have even proposed a method to adaptively weigh different auxiliary tasks.
	
	\section{Methods}
	\label{sec:methods}
	
	This section explains the neural network we use to learn representations with auxiliary tasks, the RL algorithms we train on these representations and the environments we use for training.
	
	\subsection{Representation Learning Network}
	\label{sec:ofenet}
\begin{figure}[h]
		\centering
		\includegraphics[width=0.8\textwidth]{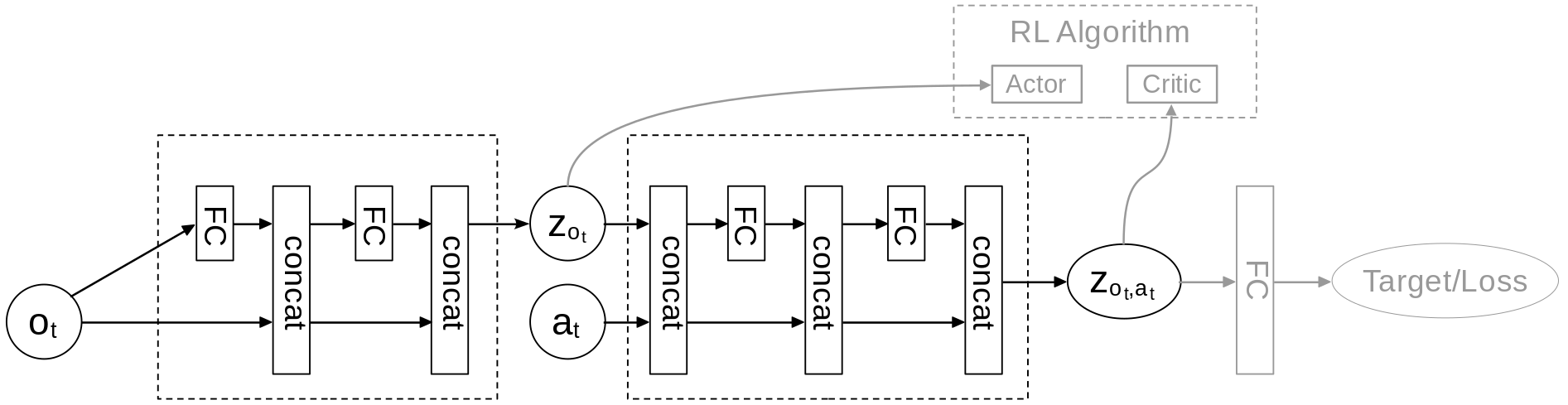}
		\caption{Sketch of the OFENet architecture, modified from \cite{ota_can_2020}. Observation $o_t$ and action $a_t$ are used to calculate representations $z_{o_t}$ and $z_{o_t, a_t}$. These are passed into the RL algorithm (light grey). The prediction target necessary to evaluate the auxiliary loss, e.g. $o_{t+1}$, is calculated with a fully connected layer (FC, light grey) from $z_{o_t, a_t}$.}
		\label{fig:ofenet_sketch}
	\end{figure}

	To train decoupled representations on auxiliary tasks, we use the network architecture of OFENet from \cite{ota_can_2020}. The architecture is composed of two parts. Its first part calculates a representation $z_{o_t}$ of $o_t$, and the second part calculates a representation $z_{o_t, a_t}$ of $z_{o_t}$ and $a_t$. Internally, the parts stack MLP-DenseNet blocks which consist of fully connected and concatenation layers. The whole arrangement is visualized in Fig. \ref{fig:ofenet_sketch}. For our experiments we give both parts of OFENet the same internal structure (apart from input dimensionality), but adjust parameters to different environments as described in Tab. \ref{tab:dimensions}. The auxiliary loss is calculated as the mean squared error between predicted and actual target.
	
	OFENet is a good choice for comparing auxiliary tasks as it is a rather generic architecture for learning representations of expanded dimensionality. Besides OFENet, we are not aware of any other decoupled approaches used in RL that learn representations without dimensionality reduction. Most works use autoencoders (variational or otherwise), which have been shown to be very powerful especially for visual data. OFENet, however, has the advantage that it can be applied to smaller, simpler environments. This allows us to study auxiliary tasks in environments that have far fewer dimensions than visual observations would have and that are less complex than those used in \cite{shelhamer_loss_2017,de_bruin_integrating_2018}.
	
	\subsection{Reinforcement Learning Algorithms}
	To solve the RL task of maximizing returns, we use TD3 \cite{fujimoto_addressing_2018} and SAC \cite{haarnoja_soft_2018}, two well-known state-of-the-art RL algorithms. They are both model-free off-policy actor-critic methods. Comparing auxiliary tasks against these two presents a trade-off between the computational expense of the runs required for our comparison (hundreds per RL algorithm) and investigating more than one algorithm to avoid results being biased. We chose these two algorithms in particular because they are powerful and also popular, which makes them a testbed that is both non-trivial and particularly relevant to readers.
	
	We study one environment, FetchSlideDense-v1, that is too difficult to solve with baseline TD3 and SAC. It does however become at least partially solvable when adding hindsight experience replay, first proposed in \cite{andrychowicz_hindsight_2017}. HER infuses the replay buffer used by off-policy algorithms with additional samples copied from previous episodes. In these copied samples it changes the reward signal to pretend the agent had performed well in order to present it with positive learning signals. Additional supposedly successful episodes provide a stronger incentive for the agent to learn, which makes learning in complex environments easier. Nowadays it is wide-spread practice to use HER for robotics tasks such as \mbox{FetchSlideDense-v1}.
	
	\subsection{Environments}

\begin{figure}[h]
		\centering
		\includegraphics[width=\textwidth]{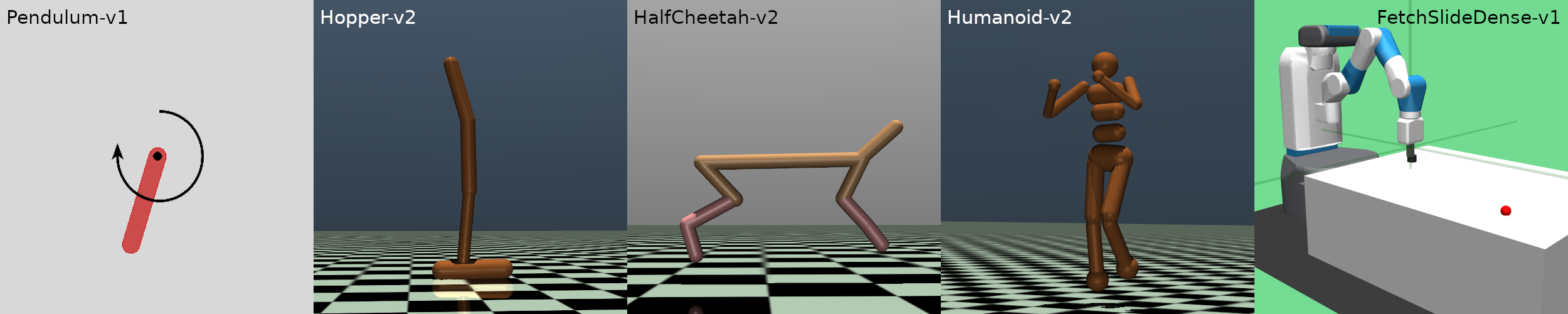}
		\caption{Sample images rendered to visualize the environments. The image of FetchSlideDense-v1 is taken from \cite{plappert_multi-goal_2018}.}
		\label{fig:env_pictures}
	\end{figure}

	We perform our study on five different environments: A simulated pendulum, three MuJoCo control tasks and a simulated robotics arm. They span a large range of size and complexity. Size, here, refers to the dimensionality of observation and action space, while complexity concerns how difficult it is to learn a sufficient mapping between observation space, action space, and rewards. The three MuJoCo control tasks differ in size but are controlled by similar dynamics, which allows for a very direct comparison. All studied environments are depicted in \mbox{Fig. \ref{fig:env_pictures}.} Sizes of observation and action spaces, and of the corresponding representations learned with OFENet, are listed in Tab. \ref{tab:dimensions}.
	
	In the following, all five environments we use are briefly described. For further details on the first four we refer the reader to OpenAI Gym's \href{https://www.gymlibrary.dev}{documentation} \cite{brockman_openai_2016}.
	
	\begin{table}[th]
		\centering
		\caption{Dimensions of observations, actions, representations, and OFENet parameters used to achieve them. Layers per part describes the total amount, and individual width, of fully connected layers per OFENet part.}
		\label{tab:dimensions}
		\begin{tabular}{l|lllll}
			\multicolumn{1}{c|}{\bf Environment}  & \multicolumn{1}{c}{\bf dim($o_t$)} & \multicolumn{1}{c}{\bf dim($a_t$)} & \multicolumn{1}{c}{\bf dim($z_{o_t}$)} & \multicolumn{1}{c}{\bf dim($z_{o_t, a_t}$)} &  \multicolumn{1}{c}{\bf Layers/part}
			\\ \hline
			Pendulum-v1 & 3 & 1 & 23 & 44 & 2 $\times$ 10\\
			Hopper-v2 & 11 & 3 & 251 & 494 & 6 $\times$ 40\\
			HalfCheetah-v2 & 17 & 6 & 257 & 503 & 8 $\times$ 30\\
			Humanoid-v2 & 292 & 17 & 532 & 789 & 8 $\times$ 30\\
			FetchSlideDense-v1 & 31 & 4 & 271 & 515 & 8 $\times$ 30
		\end{tabular}
	\end{table}
	
	\textbf{Pendulum-v1} is a simple and small classic control environment in which a pendulum needs to be swung upwards and then balanced in this position by applying torque. Its observations quantify angle and angular velocity of the pendulum. The reward at each time step is an inversely linear function of how much the angle differs from the desired goal, how much the angle changes, and how much torque is applied.
	
	\textbf{Hopper-v2} is one of three MuJoCo control tasks we consider here. It is based on a physical simulation of a two-dimensional single leg with four parts, which can be controlled by applying torque to three connecting joints. This makes it comparatively small and simple. The observation contains certain angles and positions of parts and joints, and their velocities. The reward at a given time step mostly depends on how much the hopper has moved forward, plus a constant term if it has not collapsed.
	
	\textbf{HalfCheetah-v2} is another two-dimensional MuJoCo control task, similar to Hopper-v2 but larger and more complex. It already consists of 9 links and 8 joints, with action and observation space similar in nature to those of Hopper-v2 but consequently of larger dimensionality. The reward is again based on how much the HalfCheetah-v2 has moved forward since the last time step.
	
	\textbf{Humanoid-v2} is the third MuJoCo control task we use in our comparison. As opposed to the others, it is three-dimensional. It roughly models a human, which leads to actions and observations similar to those of Hopper-v2 and HalfCheetah-v2, but of far higher dimensionality. Again, the reward is primarily based on forward movement plus a constant term if the robot has not fallen over.
	
	\textbf{FetchSlideDense-v1} is a simulated robotics task presented in \cite{plappert_multi-goal_2018}. It is not much larger than HalfCheetah-v2, but much more complex than any of the other tasks. A three-dimensional arm needs to push a puck across a low-friction table such that it slides to a randomly sampled goal position out of reach of the arm. The action controls movement of the tip of the arm, while the observation encodes position and velocities of arm and puck as well as the goal location. The reward is the negative distance between puck and goal, and thus constant until the arm hits the puck. In their technical report, the authors show that this task is very difficult to solve even with state-of-the-art methods, unless additional methods such as HER are deployed. FetchSlideDense-v1 is evaluated by success rate instead of return. Success rate describes in how many cases out of 100 the puck ended up closer than some threshold to its goal.
	
	\section{Experiments}
	To compare the auxiliary tasks, we train agents with baseline TD3 and SAC on raw observations (baseline) and on representations learned with auxiliary tasks. We do this for each of the five environments. Additionally, for FetchSlideDense-v1, we combine TD3 and SAC with HER and train these on raw observations as well as on representations learned with auxiliary tasks. All of the aforementioned experiments are conducted five times with the same set of random seeds. We do regular evaluations over several evaluation episodes throughout training, and their average return/success rate is what we report here.
	
	For our experiments we use the PyTorch implementations of TD3 \cite{fujimoto_addressing_2022} and SAC \cite{yarats_soft_2022}, together with our own PyTorch implementation of OFENet based on the Tensorflow code provided with \cite{ota_can_2020}. For the experiments with HER, we modified the Stable-Baselines3 code \cite{raffin_stable-baselines3_2021} to include OFENet. For the experiments done with Stable-Baselines3, we took hyperparameters from the RL Baselines3 Zoo repository \cite{raffin_rl_2020}. In all other experiments, hyperparameters are the default ones provided by the respective RL algorithm implementation or the OFENet implementation of \cite{ota_can_2020}, configured as indicated in Tab. \ref{tab:dimensions}.
	
	We pretrain OFENet with 1000 steps for Pendulum, and 10,000 steps for the other environments. This pretraining data is sampled using a random policy. After that, the system alternates between training OFENet on its auxiliary task and the RL algorithm on its RL task, while freezing the weights of the respective other. Representations are thus continuously updated during the training process and become optimized on those states and actions relevant to the agent. In each iteration OFENet and agent are trained on the same sampled observations and we count this as one training step of the overall system. In other words, we only count training steps of the RL algorithm for better clarity and comparability.
	
	In terms of computation time, adding OFENet to the RL algorithms roughly doubles to triples the training time of our agents, which appears little given the large increase in dimensionality. We speculate that this factor is caused by a doubling in backward passes for gradient updates plus some additional overhead in handling two separate networks for separate tasks, while additional gradients due to the increased network width can be computed in parallel by PyTorch.
	
	\section{Results}
	\label{sec:results}
	
\begin{figure}[h]
		\centering
		\includesvg[width=\textwidth]{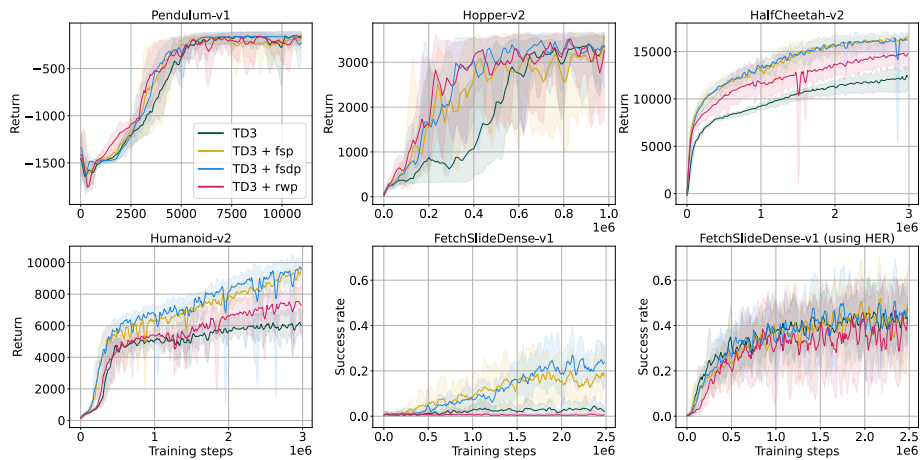}
		\caption{Returns/success rates achieved with TD3 and different auxiliary tasks on various environments. The shaded areas show minimum and maximum performance achieved across 5 runs, while the lines represent the means. Values have been smoothed slightly for better visualisation.}
		\label{fig:returns_td3}
	\end{figure}

\begin{figure}[h]
		\centering
		\includesvg[width=\textwidth]{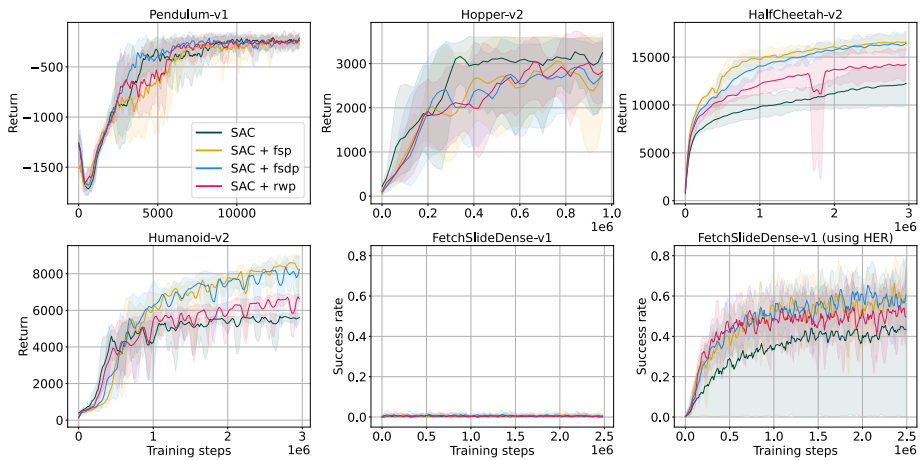}
		\caption{Returns/success rates achieved with SAC and different auxiliary tasks on various environments. The shaded areas show minimum and maximum performance achieved across 5 runs, while the lines represent the means. Values have been smoothed slightly for better visualisation.}
		\label{fig:returns_sac}
	\end{figure}

\begin{figure}[h]
		\centering
		\includesvg[width=\textwidth]{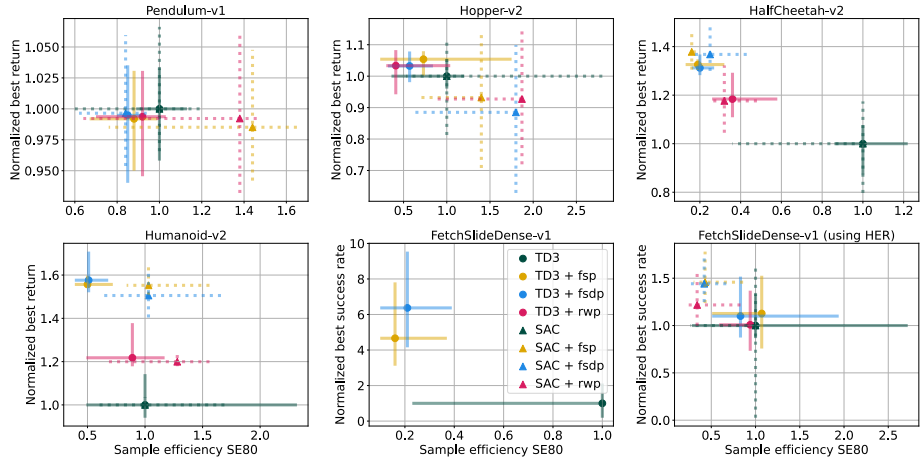}
		\caption{Sample efficiency on different environments compared to normalized best returns/success rates. Note the different scales on the axes. The markers describe average performance, error bars (solid for TD3 and dotted for SAC) mark best and worst case out of 5 runs. Where markers or error bars are missing, agents in question never reached the return/success rate required to calculate SE80.}
		\label{fig:maxr_se}
	\end{figure}

	The returns or success rates on all different environments are shown in Figs. \ref{fig:returns_td3} and \ref{fig:returns_sac} for all auxiliary tasks and baseline algorithms. For a direct, normalized comparison Fig. \ref{fig:maxr_se} plots the normalized maximum return/success rate against sample efficiency. To measure sample efficiency, we calculate the fraction of training steps (and therefore samples) which are required to reach 80\% of the maximum return of the baseline algorithm, calibrated against the untrained baseline algorithm since the initial reward is not always 0. We choose 80\% instead of 100\% because at this lower threshold we can capture increases in sample efficiency even where maximum return/success rate are similar to that of the baseline. We call our measure SE80.
	
	
	In the following, the word \textit{performance} shall refer to the combination of maximum return and sample efficiency. If only one of the two is concerned, we will state that explicitly. It is apparent that all three auxiliary tasks lead to a significant increase in performance for complex, higher-dimensional environments. For the low-dimensional and simple Pendulum-v1, a slight increase in sample efficiency but not in best return can be achieved. In fact, improvements in sample efficiency are achieved across almost all environments. Increases in maximum returns follow a certain pattern: They seem to increase with problem complexity rather than strictly dimensionality, although the two go hand in hand. When using HER to solve FetchSlideDense-v1, however, representation learning only leads to minor improvements. This is a special case discussed in \mbox{Sect. \ref{sec:comp_envs}}.

	\subsection{Representation Learning for Different Types of Environments}
	\label{sec:comp_envs}
	Our experiments show different behavior for small and simple environments compared to larger and more complex ones.
	For the very small and simple Pendulum-v1 environment, representation learning with auxiliary tasks does not significantly benefit return or sample efficiency. For the slightly larger and less simple Hopper-v2 environment, the picture is ambiguous with an increase in sample efficiency for TD3 but not for SAC. For the remaining larger and more complex environments, however, representation learning with auxiliary tasks provides clear performance gains over baseline TD3 and SAC. These gains seem to scale with complexity rather than size of the environments, as the difference in performance between HalfCheetah-v2, FetchSlideDense-v1 and Humanoid-v2 is not proportionate to their difference in size.
	
	An interesting case is the FetchSlideDense-v1 environment. It is too complex for any learning to occur with baseline TD3 or SAC (without HER). Because of its initially constant rewards, \textit{rwp} is not able to learn anything at all. Adding HER to the RL algorithm, however, seems to speed up learning enough to generate meaningful rather than trivial reward signals very soon and to successfully train \textit{rwp}, as evidenced by the fact that agents using \textit{rwp} are competitive with those trained on other tasks.
	
	The authors proposing HER argue that in cases such as FetchSlideDense-v1 too few learning impulses, in the form of rewards, are provided to meaningfully update network weights in the RL algorithm. When using TD3 with HER, the auxiliary tasks do not seem to offer any benefits. For SAC with HER, the agents trained with auxiliary tasks are on average better than those without. Baseline SAC with HER can in fact perform as well as with auxiliary tasks, but is less reliable; its mean is lowered considerably by two agents which did not learn at all. These results suggest that adding a learning signal through HER in principle enables the RL algorithm itself to learn meaningful patterns from original observations (i.e. HER significantly reduces the complexity of the problem), but it only does so reliably when adding representations learned on auxiliary tasks.
	
	Furthermore, FetchSlideDense-v1 becomes at least partially solvable for TD3, even without HER, when \textit{fsp} or \textit{fsdp} are used. This interesting result shows that even if an environment is too complex for a RL algorithm, adding representation learning might still make training of agents possible. There is however no equal improvement in the same experiment with SAC, which shows that this strategy has its limits. We hypothesize that the learned representations recast observations, actions and thereby the entire RL problem into a less complex manifold. At least some dimensions of the representation learned with OFENet would then contain more informative features than the original observation. For \mbox{FetchSlideDense-v1} the representation might for instance contain a feature encoding distance between arm and puck, instead of just the absolute positions from raw observations. When the arm accidentally hits the puck, the RL algorithm could consequently relate observation and reward more easily.
	Another possible factor, proposed in \cite{ota_can_2020}, is that the added depth and width of OFENet enable the agent to learn more complex and therefore more successful solutions. In this case, however, additional expressivity through added weights alone does not reduce problem complexity which is caused by initially constant rewards. It can therefore not explain why \textit{fsp} and \textit{fsdp} make FetchSlideDense-v1 learnable for TD3 without HER. We thus consider simplification of the learning problem to be the dominant factor at least for this setting.
	
	\subsection{Comparison of Auxiliary Tasks}
	
	This section presents a direct comparison of auxiliary tasks across the different algorithms and environments. Since HER seems to significantly distort the performance of auxiliary tasks compared to just using baseline RL algorithms, the FetchSlideDense-v1 solved with HER will not be considered.
	
	In the remaining cases, the \textit{rwp} task performs worst out of all investigated tasks. For the complex and high-dimensional environments it is quickly outperformed by \textit{fsp} and \textit{fsdp}, even though it appears competitive for environments with less complex dynamics where differences in performance are minimal. The performances of \textit{fsp} and \textit{fsdp} are approximately similar, although one usually outperforms the other by a slight but noteworthy margin. There is however no apparent pattern to this. When used with TD3, there might be a slight tendency for \textit{fsdp} to outperform \textit{fsp}, but results are too inconclusive to confidently make this claim, especially since it cannot be observed with SAC-based agents.
	
	There are three potential causal factors which might explain why \textit{rwp} performs worse. Firstly, due to its dimensionality alone, the prediction target $r_{t+1}$ of \textit{rwp} can not convey the same amount of information as the prediction targets of \textit{fsp} and \textit{fsdp}. Secondly, the nature of the information differs. Learning representations on environment dynamics makes environment information accessible that is much harder to access when using reward signals. Thirdly, the reward signals are provided to both the agent and OFENet, which underlines the redundancy claim regarding \textit{rwp}. However, neither of these factors is easy to investigate without studying the representations. We hope to conduct such an investigation as future work to better understand these explanations.
	
	The absence of a consistent difference in performance between \textit{fsp} and \textit{fsdp} suggests that the theoretical advantages of each (Sect. \ref{sec:aux_tasks}) are either not important or cancel each other out. Our studied environments are well behaved as they all simulate real-world physics. Consequently, they do not confront the algorithm with abrupt state changes or excessive noise. The fact that \textit{fsp} on average still works about as well as \textit{fsdp}, despite those properties, suggests that the advantages proposed for \textit{fsdp} in particular do not play a large role in practice.
	
	\section{Conclusion}
	
	In this paper we compare auxiliary tasks for decoupled representation learning on non-visual observations in RL. To this end we use five common benchmark environments and two different state-of-the-art off-policy RL algorithms. We find that representation learning with auxiliary tasks can significantly improve both sample efficiency and maximum returns for larger and more complex environments while it makes little difference with simpler environments that are easy to solve for the baseline agent. In those latter cases, we observe a slight increase in sample efficiency at most. Auxiliary tasks that encourage learning environment dynamics generally outperform reward prediction. Particularly encouraging is that the FetchSlideDense-v1 environment, a simulated robotics arm, becomes partially solvable when adding representation learning to the otherwise unsuccessful TD3 algorithm. We interpret this as an indication that decoupled representation learning with auxiliary tasks can reduce problem complexity in RL. Across all experiments, we found that results might vary between RL algorithms, even when using the same representation learning techniques.
	
	Despite this variability we are confident that the patterns we found can contribute to the future development of representation learning algorithms for RL, in particular for decoupled and interpretable representation learning approaches for real-world applications.

	\bibliographystyle{splncs04}
	\bibliography{paper}
	
\end{document}